# Unsupervised Ensemble Ranking of Terms in Electronic Health Record Notes Based on Their Importance to Patients

Jinying Chen[1] and Hong Yu[2,1]

[1]Department of Quantitative Health Sciences, University of Massachusetts Medical School, Worcester, MA, USA

[2] Bedford Veterans Affairs Medical Center, Center for Healthcare Organization and Implementation Research, Bedford, MA, United States

## Abstract

**Background:** Allowing patients to access their own electronic health record (EHR) notes through online patient portals has the potential to improve patient-centered care. However, EHR notes contain abundant medical jargon that can be difficult for patients to comprehend. One way to help patients is to reduce information overload and help them focus on medical terms that matter most to them. Targeted education can then be developed to improve patient EHR comprehension and the quality of care.

**Objective:** The aim of this work was to develop FIT (Finding Important Terms for patients), an unsupervised natural language processing (NLP) system that ranks medical terms in EHR notes based on their importance to patients.

**Methods:** We built FIT on a new unsupervised ensemble ranking model derived from the biased random walk algorithm to combine heterogeneous information resources for ranking candidate terms from each EHR note. Specifically, FIT integrates four single views (rankers) for term importance: patient use of medical concepts, document-level term salience, word co-occurrence based term relatedness, and topic coherence. It also incorporates partial information of term importance as conveyed by terms' unfamiliarity levels and semantic types. We evaluated FIT on 90 expert-annotated EHR notes and used the four single-view rankers as baselines. In addition, we implemented three benchmark unsupervised ensemble ranking methods as strong baselines.

**Results:** FIT achieved 0.885 AUC-ROC for ranking candidate terms from EHR notes to identify important terms. When including term identification, the performance of FIT for identifying important terms from EHR notes was 0.813 AUC-ROC. Both performance scores significantly exceeded the corresponding scores from the four single rankers ($P$<.001). FIT also outperformed the three ensemble rankers for most metrics. Its performance is relatively insensitive to its parameter.

**Conclusions:** FIT can automatically identify EHR terms important to patients. It may help develop future interventions to improve quality of care. By using unsupervised learning as well as a robust and flexible framework for information fusion, FIT can be readily applied to other domains and applications.

## 1. Introduction

Online patient portals have been widely adopted by health systems in the U.S. in a nationwide effort to promote patient-centered care [1–3]. In addition, the *OpenNotes* initiative [4] and the Blue Button movement [5] allow patients to access their full EHR notes through patient portals, with early evidence showing improved medical comprehension, healthcare management, and outcomes [6–8].

However, patients may face great challenges when reading full EHR notes due to the lack of medical training. EHR notes are typically long and contain abundant medical jargon. Previous studies showed that EHRs were written at an $8^{th}$-$12^{th}$ grade reading level [9–12], which is above the average adult patient's reading level of $7^{th}$ to $8^{th}$ grade in the U.S. [13–17]. In addition, 36% adult Americans have limited health literacy [18]. Limited health literacy has been identified as one of the major barriers to patient portal use (which includes interpreting information from EHRs) [19–21].

One way to help patients to comprehend their EHR notes is to reduce information overload and help them first understand medical terms that matter most to them. This approach is motivated by two reasons. First, medical terms have been shown to be obstacles for patients [22–27]. Second, EHR notes incorporate a comprehensive description of patients' medical courses, part of which (e.g., technical details about surgery procedures or echocardiogram results) may not directly address patients' immediate concerns. The approach of explaining all the jargon in their notes at once may likely overwhelm the patients and may be unnecessary in the first place. Therefore, we help patients focus on EHR terms most important to them. Personalized interventions can then be developed to support patient EHR comprehension, for example, using important terms identified from a patient's EHR note to retrieve educational materials targeting on this patient.

Figure 1 shows an excerpt from a typical EHR note from our evaluation data which was annotated by physicians. Although there are many medical terms in this piece of text (here we only highlighted a subset of terms identified by the Unified Medical Language System [UMLS] lexical tool MetaMap [28] for illustration purpose), physicians identified only five terms most important for patients to know, i.e., "pancreatic neoplasm", "Whipple procedure", "pancreatectomy", "splenectomy", and "insulin-dependent diabetic". Physicians judged a term's importance based on whether the patient should know this term in order to better understand the most important aspects medically relevant to his/her health and treatment course. Note that physicians do not mark many unfamiliar medical terms, e.g., "Tinnitus", "CA 19-9", and "bile duct stricture", suggesting that they do not rank terms based on their difficulty levels.

**Figure 1.** A sample EHR text where physicians identified important medical terms (underlined). Other medical terms are italicized.

> A lovely xx-year-old gentleman with multiple issues for comprehensive evaluation.
>
> 1. *Tinnitus* of many years' duration. He has no *neurologic complaints*, no headache, no *vertigo*, no sudden changes in hearing.
>
> 2. Low-grade <u>*pancreatic neoplasm*</u>.
> In xxxx, presented with *epigastric pain* and high *CA 19-9*. *CT* showed a mass in the *pancreas*.
>
> He underwent a <u>*Whipple procedure*</u> with a <u>*pancreatectomy*</u> and <u>*splenectomy*</u> showing a *mucinous neoplasm* with *secondary pancreatitis*.
>
> This was complicated by a *bile duct stricture* requiring reconstruction in xxxx.
>
> In xxxx, he had *HIB*, *Meningovax* and *Pneumovax*, and he had a flu shot this fall.
>
> In xxxx, he had some *epigastric pain* prompting a contrast *CT of the abdomen* and *pelvis* that did not show any *tumor recurrence*.
>
> The patient is on *pancreatic enzyme* replacement. He is also an <u>*insulin-dependent diabetic*</u>.
>
> He tells me he had one bout of abdominal pain that may be xxxx that lasted about 4 hours in low abdomen, crampy without diarrhea or vomiting and it has gone away. He has had nothing since. No pain, burning or blood with urination. Appetite is excellent. There is no *melena* or bad diarrhea.

The goal of this work is to develop a robust unsupervised NLP system, called Finding Important Terms for patients (FIT), to automate the process of identifying EHR(patient)-specific important terms. This task is challenging because FIT does not use supervision from labeled data. In addition, existing unsupervised methods based on general principles of term importance (such as term frequency and topic coherence) are not sufficient to solve this problem (details in the Discussion section). We address this challenge by proposing a new unsupervised ensemble ranking method which adapts the biased random walk algorithm for information fusion to integrate evidences of term importance from heterogeneous information resources. We empirically show that, using this method, FIT outperforms other state-of-the-art unsupervised ensemble models on the task of ranking EHR terms.

## 2. Related Work

### 2.1. NLP Systems Facilitating Concept-level EHR Comprehension

There has been active research on linking medical terms to lay terms [10,29,30] and also on linking them to consumer-oriented definitions [11] and educational materials [31], and showing improved comprehension with such interventions [10,11].

On the issue of determining which medical terms to simplify, previous work explored frequency-based and/or context-based approaches to check if a term is unfamiliar to the average patients or if it has simpler synonyms [10,29,30]. Such work targets medical jargon and treats all the jargon terms equally important. In our work, we used term unfamiliarity as partial evidence for term importance.

It is worth noting that our approach is complementary to previous work. For example, in a web-based system supporting EHR comprehension such as [11], we can display lay definitions for all the medical jargon in a patient's EHR note, and then highlight those terms which FIT predicts to be most important to this patient by using background color and also link them to educational materials.

Our recent work shows that a supervised learning-to-rank system trained on in-domain data is effective in identifying important terms from EHR notes [32]. The work we present here studies unsupervised methods for better domain portability, because they can be easily applied to different domains without using manually annotated training data.

### 2.2. Unsupervised Single-Document Keyphrase Extraction

Our work is related to but different from single-document keyphrase extraction (KE), which identifies terms representing important concepts and topics in a document. KE targets topics which the writers wanted to convey when writing the documents. Our problem is more challenging because physicians wrote EHR notes for physician-physician communication and, therefore, features extracted from EHR notes may not be sufficient to guide an automated system to find topics and terms important to patients.

Previous work in unsupervised KE has explored various techniques, including language modeling, topic-clustering, graph-based ranking and simultaneous learning of keyphrases and key sentences [33]. Among them, graph-based methods such as TextRank [34] and its variations are the state-of-the-arts [33]. We adapted SingleRank [35] (an extension of TextRank) to clinical domain and used it as a baseline as well as an input for the ensemble ranking approaches.

KE in the biomedical domain has been limitedly explored in literature articles and in using domain-specific methods and features [36,37]. For example, Li and Wu [36] developed KIP to extract keyphrases from medical articles. KIP used MeSH (Medical Subject Headings) as the knowledge base to compute a score to reflect a phrase's domain specificity. It assigned each candidate phrase a rank score by multiplying its within-document term frequency and domain-specificity score. Song and Tanapaisankit [37] developed BioKeySpotter based on the assumption that keyphrases in biomedical articles are either biomedical entities or their neighboring terms. They used biomedical entity detection and dependency parsing to extract candidate terms and used mutual information to estimate term importance. In this work, we use the consumer health vocabulary (CHV) as a knowledge resource to extract information about term importance. We assume that important EHR terms are a subset of the terms extracted by MetaMap [28].

**Figure 2.** Overview of FIT and its evaluation.

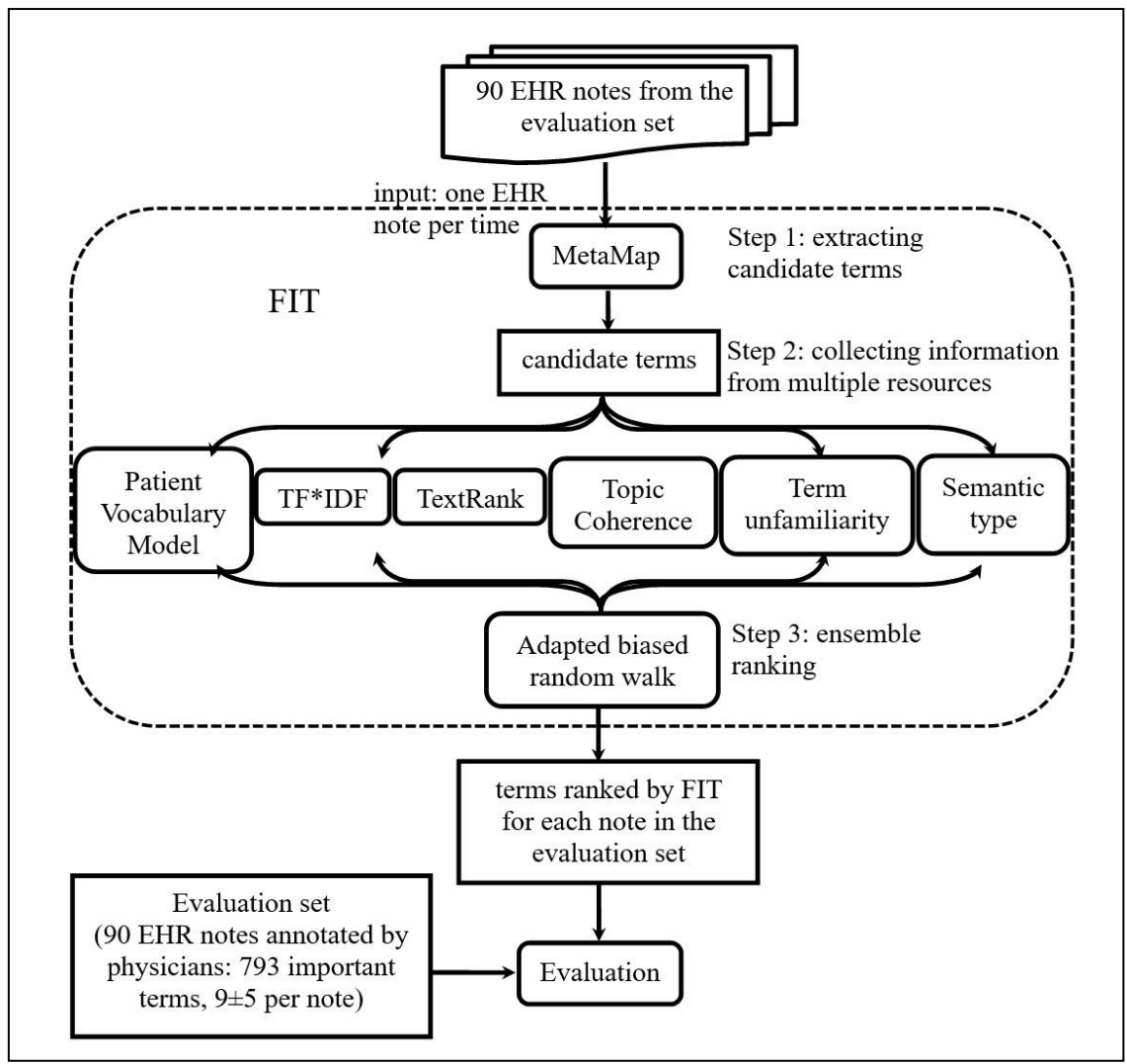

## 3. The FIT System

Figure 2 shows the workflow of running and evaluating FIT. FIT operates in three steps: identifies candidate terms, collects information about term importance from multiple resources, and ranks terms. In the first step, FIT applies MetaMap [28], a concept detection tool that automatically maps biomedical text to UMLS concepts, to find medical terms as candidate terms. The remaining pipeline of FIT is detailed below.

### 3.1 Four Single Views of Term Importance

Unsupervised ensemble ranking methods typically assume that the single rankers to combine have two properties [38,39]: (1) goodness—the outputs from the single rankers have reasonably good qualities; and (2) diversity—the single rankers are sufficiently different and complimentary to each other. We follow these principles when designing the single rankers or views of term importance.

#### 3.1.1 Patient Use of Medical Concepts

In this view, we consider medical terms frequently used by patients to be important to patients (w.r.t. patient centeredness). One way to quantify terms used by patients is to collect and analyze patients' queries on the internet.

The consumer health vocabulary (CHV) is collection of terms used by patients. It is a rich resource that incorporates terms extracted from various consumer health sites, such as queries submitted to MedLinePlus and postings in health-focused online discussion forums [38–44]. It contains 152,338 terms, most of which are consumer health terms [42–44]. Zeng et al. [43] mapped these consumer health terms to the UMLS concepts by a semi-automatic approach. As the result of this work, the CHV encompasses lay terms as well as corresponding medical terms.

Although the CHV is a comprehensive patient-centered resource, its coverage is limited. Therefore, we used the CHV to develop a distant-supervision vocabulary model to rank the patient-centeredness of a medical term.

Our model made an assumption that medical terms that occur in both EHRs and the CHV are important to patients for comprehending their EHRs because they are medical synonyms of terms initially used by patients to query online health forums. We built the model on support vector machines (SVMs). The training examples include 37,732 terms identified from 6K EHR notes by MetaMap. We followed [29] (i.e., CHV familiarity score $\leq 0.6$) to identify 9,994 medical jargon terms that occur in the CHV and also belong to these EHR terms, which we labeled as positive examples. The remaining 27,738 terms were labeled as negative.

We used word embeddings as learning features because word embedding has emerged as a powerful technique for word representation and has been successfully used in biomedical and clinical NLP tasks such as biomedical named entity recognition [45,46], adverse drug event detection [47,48], ranking biomedical synonyms [49], and disambiguating clinical abbreviations [50,51].

We trained a neural language model to learn word embeddings by using the Word2Vec software [52,53]. We trained Word2Vec on a combined text corpus (over 3G words) of English Wikipedia, articles from PubMed Open Access and 99K EHR notes from the Pittsburg corpus[1]. We set the training parameters based on the study of Pyysalo et al. [54]. We used 200-dimension word vectors learned by Word2Vec as features to learn the vocabulary model. For a multi-word term, we used the mean of its component words' vectors as features by following [49]. We normalized each dimension of the feature vectors to [0,1] by min-max scaling.

We built the vocabulary model on the RBF-kernel SVM algorithm implemented by LibSVM [55]. We optimized model parameters on the training data and used the trained model to classify the candidate terms into two categories: patient-centered medical terms (as represented by medical terms that occur in both the CHV and EHRs) and other terms. Given test instances (i.e., candidate terms), SVM's original outputs are the signed distances between these instances and the classification decision boundary. LibSVM transforms these distance values into probabilities by Platt scaling [56,57]. We used the respective probabilities of each candidate term being a positive instance to rank candidate terms from each EHR note.

### 3.1.2 Document-level Term Salience

FIT uses TF*IDF to represent the salience of a candidate term to an individual EHR note. TF*IDF [58] is widely used to measure the salience of a term to a document $d$ in a corpus $D$, as defined in (1). The more frequent the term appears in the document and the less frequent it appears in other documents, the more important it is to this document.

$$TF * IDF(t, d, D) = TF(t, d) * IDF(t, D)$$
$$IDF(t, D) = log \frac{N}{|\{d | t \in d\}|} \quad (1)$$

Here, $t$ is a term; $d$ is a document; $TF(t, d)$ is the frequency of $t$ in $d$; $IDF(t, D)$ is the inverse document frequency of $t$ in corpus $D$; $N$ is the total number of documents in corpus $D$. We used as $D$ the 6K EHR notes collected from the same domains where the evaluation data was collected.

### 3.1.3 Word co-occurrence Based Term Relatedness

---

[1] Chapman W., University of Pittsburgh NLP Repository. Using this data requires a license.

We used SingleRank [35] to represent this view. In our case, each EHR note is an undirected, unweighted graph in which words are vertices and are connected if they co-occur within a context window of 10 [35]. The connected words are treated as neighbors of each other. The rank score of a word is calculated recursively by (2),

$$S(v_i) = (1-d) \times \frac{1}{N} + d \times \sum_{v_j \in Adj(v_i)} \left( \frac{\omega_{ji}}{\sum_{v_k \in Adj(v_j)} \omega_{jk}} S(v_j) \right) \quad (2)$$

where $v_i$ is a word, $Adj(v_i)$ is a set that contains all neighbors of $v_i$, $d$ is the damping factor which is set to 0.85 [59], $\omega_{ji}$ is the edge weight which equals the number of co-occurrences of $v_i$ and $v_j$ within a context window of 10. $\frac{\omega_{ji}}{\sum_{v_k \in Adj(v_j)} \omega_{jk}}$ is the probability of reaching $v_i$ by $v_j$.

The rank score of a candidate term is the sum of the rank scores of individual words contained in this term.

### 3.1.4 Topic Coherence

In this view, the importance of a candidate term to an EHR note is measured by the topic coherence between the term and the note. We compute topic coherence $P(t|e)$ by (3) and (4),

$$P(w|e) = \sum_{i=1}^{K} P(w|topic_i) P(topic_i|e) \quad (3)$$

$$P(t|e) = \sum_{w \in t} P(w|e) \quad (4)$$

where *P(t|e)* is the probability of a candidate term conditioned on an EHR note *e*; *P(w|e)* is the probability of a word *w* conditioned on *e*; $P(w|topic_i)$ and $P(topic_i|e)$ are word-topic and topic-EHR note distributions estimated by the topic model; *K* is the number of topics used in topic modeling.

We used the Latent Dirichlet Allocation algorithm implemented by MALLET [60] for topic modeling. We trained the topic model on the same 6K EHR notes which were used to compute TF*IDF and set *K* to 200 after testing different *K*'s on the training data.

### 3.2 Additional Information about Term Importance

### 3.2.1 Term Unfamiliarity

Familiar terms (such as "flu" and "cough") are likely to be already known by patients and therefore may not be important for interventions that support patient's EHR comprehension. Previous work adopts this assumption and simplifies only difficult terms [10,11,29,30]. We follow this idea and assume that, although an unfamiliar term is not necessarily important to a patient, unfamiliar terms are in general more important and thus should be ranked higher than familiar ones.

We used CHV familiarity scores to distinguish between unfamiliar terms and familiar terms. The CHV assigns familiarity scores to 58% (88,189 out of 152,338) of its terms for extended usability. CHV familiarity scores estimate the likelihood that a medical term can be understood by an average reader [61] and take values between 0 and 1 (with 1 being most familiar and 0 being least familiar). The CHV provides different types of familiarity scores [29]. Following [29], we used the combined score and a score threshold 0.6 to detect unfamiliar and familiar terms (familiarity score ≤0.6 are unfamiliar terms; >0.6 are familiar terms). Since not every EHR term has a CHV familiarity score, the available information about term unfamiliarity is partial.

Note that we divide CHV terms into only two categories (unfamiliar vs. familiar terms) rather than rank them by fine-grained term familiarity scores, because the most unfamiliar terms (with the lowest familiarity score in the CHV) are not necessarily the most important terms.

*3.2.2 Semantic Types of Medical Concepts*

As introduced in section 3.1.1, CHV terms have been mapped to UMLS concepts [43] and thus have UMLS semantic types (in total, 134 semantic types in the CHV). We define the frequency of a semantic type in the CHV as the number of medical concepts of this type in the CHV. We found that the frequency distribution of semantic type in the CHV is highly skewed (see Figure A.1 in Appendix A). This suggests that medical concepts frequently used by patients may concentrate on a small number of semantic types (which we called CHV-preferred semantic types). We therefore made an assumption that medical terms with CHV-preferred semantic types are more important to patients than terms with other semantic types and thus should be ranked higher.

Since the frequency distribution of semantic type in the CHV is highly skewed with a long tail, we ranked the semantic types by frequency and identified the cutting point using two criteria: (1) altogether, the semantic types above the cutting point cover over 50% of the medical concepts in the CHV; and (2) among all candidate cutting points that satisfy (1), the cutting point represents the largest frequency gap between adjacent ranks (i.e., between the lowest selected rank and the highest unselected rank). Using this method, we found the cutting point frequency>1,000 and selected 12 most frequent semantic types as CHV-preferred, which cover over 60% of the medical concepts in the CHV. The CHV-preferred types (see Appendix B for the full list) include "Pharmacologic substance" (e.g., "Budesonide"), "Disease or syndrome" (e.g., "autoimmune hemolytic anemia"), "Finding" (e.g., "retinopathy"), etc. The non-CHV-preferred semantic types include "Bacterium" (e.g., "E. coli"), "Body function" (e.g., "endocrine"), "Individual behavior" (e.g., "tobacco cessation"), "Manufactured object" (e.g., "treadmill"), etc.

Familiar and unfamiliar terms can have the same semantic type. For example, both "flu" and "pancytopenia" are assigned the semantic type "Disease and syndrome". In addition, non-CHV terms can have CHV-preferred semantic types. For example, "livedoid vasculopathy", although not a CHV term, has the CHV-preferred type "Disease and syndrome".

Note that the evidence of term importance from CHV-preferred semantic types is coarse-grained because it only divides candidate terms into two groups: likely important and unlikely important, according to their semantic types.

### 3.3 Combining Heterogeneous Information Resources by Random Walk

Our goal is to rank EHR terms based on their importance to patients. As described previously, we have collected information from multiple resources which represent term importance from different perspectives. However, we do not know which perspective is more relevant to the patient perspective. Since these information resources are complimentary to each other, an ensemble model that utilizes all of them is expected to be more robust than any model that uses only a single resource.

We used unsupervised ensemble learning and built FIT on an adapted random walk algorithm derived from PageRank [59]. The adaption is to allow transition from term $t_j$ to term $t_i$ during random walk *if and only if* term $t_j$ is ranked lower than term $t_i$ based on comprehensive information from all available resources (i.e., single rankers).

Specifically, for each EHR note, we generate a directed graph where candidate terms from the note are treated as vertices. We form an edge from term $t_j$ to term $t_i$ *if and only if* both the rank relation $R(t_j \to t_i)$ defined in (5) and the edge weight defined in (6) are greater than 0.

$$R(t_j \to t_i) = |\{r|r(t_i) < r(t_j)\}| - |\{r|r(t_j) < r(t_i)\}| \tag{5}$$

$$\omega_{j,i} = R(t_j \to t_i) \times \sum_r (\frac{1}{r(t_i)} - \frac{1}{r(t_j)}) \tag{6}$$

Here, $r$ is any single ranker that ranks both $t_i$ and $t_j$; $r(t)$ is the rank assigned to term $t$ by ranker $r$ (a small value of $r(t)$ represents a high rank); $|\{r|r(t_i) < r(t_j)\}|$ is the number of single rankers that rank $t_i$ higher than $t_j$.

Because some candidate terms do not have familiarity scores, the single ranker that uses this information cannot rank all of the candidate terms. We estimated *r(t)* for this ranker by using all the candidate terms that have familiarity scores.

Our random walk algorithm estimates the importance of each term by counting the support from its neighboring terms (i.e., any term that has an edge pointing to this term) recursively. Neighbors of higher importance and with higher probabilities to reach a term contribute more to the term's importance. Mathematically, we used the same equation as defined in (2) to update the importance score of each term, where the word $v_i$ ($v_j$) is replaced by the term $t_i$ ($t_j$).

The aforementioned algorithm uses only rank orders. Rank scores, when available, are also useful for ensemble ranking [62,63]. We therefore extended our model, by using the framework of a biased random walk model, to incorporate rank scores, as defined in (7),

$$S(t_i) = (1-d) \times p_i + d \times \sum_{v_j \in Adj(t_i)} \left( \frac{\omega_{ji}}{\sum_{v_k \in Adj(t_j)} \omega_{jk}} S(t_j) \right) \qquad (7)$$

where we replace the constant $1/N$ in (2) by $p_i$, the probability of the random jump to $t_i$. We calculate $p_i$ by (8),

$$p_i = \frac{1}{Z} \times \sum_r s_r(t_i) \qquad (8)$$

where $Z$ is a normalization constant that ensures $\sum_{i=1}^{N} p_i = 1$; $r$ is a single ranker that outputs rank scores for all candidate terms; $s_r(t_i)$ is $t_i$'s normalized rank score as assigned by $r$. We use the standard zero-one normalization method (i.e., $\frac{score_{max} - score}{score_{max} - score_{min}}$) to normalize rank scores. Note that the parameter $d$ in (7) serves a different purpose than the original damping factor and is used to weight the contributions of rank orders and rank scores to the ranking. We set $d$ to 0.5.

### 4. Experimental Settings

### 4.1 Evaluation Set

Our evaluation set contains 90 de-identified physician-annotated EHR notes from our previous work [32]. To maximize the representativeness, we selected notes from patients with six different but common primary clinical diagnoses: cancer, chronic obstructive pulmonary disease, diabetes, heart failure, hypertension, and liver failure. For each note, we asked physicians to identify at least 5 most important medical terms which the patients should know in order to better understand the most important aspects medically relevant to their health and treatment courses. We used expert annotations because this task requires a full comprehension of EHR notes which is beyond the capacity of the average patients [10–12,29]. We developed an annotation guideline (see Appendix C) to instruct physicians to annotate the notes from the patient perspective. For each note, we obtained annotations from two physicians and used the agreement from both physicians as the gold-standard. The annotation agreement (micro average) on the 90 notes is 0.51 Cohen's Kappa.

In total, the physicians have identified 793 important medical terms (9±5 terms per note), which covers a wide range of topics including disease, syndrome, medication, and diagnostic and therapeutic procedures. Table 1 summarizes the statistics of the evaluation set. We found that physicians often selected diseases and other information (e.g., stage evaluation and treatment plan) that are of immediate concern to patients, thus excluding other comorbidity diseases, for example. On the other hand, we point out that patients' information needs may differ from physicians' perception [64] and therefore important medical terms judged by physicians may not be the same as judged by patients. We will evaluate such a discordance in our future work.

Table 1. Statistics of the evaluation set.

|  | Evaluation set |
| --- | --- |
| Number of notes | 90 |
| Number of words per EHR note (mean ± std) | 816±133 |
| Number of candidate terms identified by MetaMap per EHR note (mean ± std) | 250±42 |
| Number of important medical terms identified by physicians per EHR note (mean ± std) | 9±5 |

This dataset, although a small size, represents a random sample of the larger EHR data because our NLP systems are unsupervised and do not "see" the data (i.e., were not trained by examples from this data). In addition, previous work has shown that 60-100 documents are sufficient to evaluate systems performing similar tasks [32,65].

### 4.2 Baseline Systems

We used four single-view rankers (details in section 3.1), called Patient Vocabulary Model, TF*IDF, Adapted SingleRank, and Topic Coherence respectively, as baseline systems to test the effect of ensemble ranking.

In addition, we implemented three benchmark unsupervised ensemble ranking methods, CombSum [62], Condorcet-fuse [66], and Reciprocal rank fusion [67], as strong baselines. The three methods have been widely in information retrieval and NLP, including document retrieval [62,66–69], web blog retrieval [70], opinion extraction [71], summarization [72], and entity linking [63].

### 4.2.1 CombSum

CombSum[62] is a rank-score-based ensemble method, which calculates the rank score of a candidate term $t$ by summing $t$'s rank scores received from single rankers, as calculated by (9),

$$CombSum\,(t) = \sum_{\{r|r \in R\}} s_r(t) \qquad (9)$$

where $R$ is the set of single rankers to ensemble; $s_r(t)$ is the rank score of $t$ given by a single ranker $r$.

### 4.2.2 Condorcet Fuse

Condorcet Fuse[66] sorts candidate terms by pairwise rank relation $R(t_j \to t_i)$ as defined in (5). Specifically, it ranks $t_i$ higher than $t_j$ if $R(t_j \to t_i) > 0$. We implemented the Condorcet Fuse ranker using the quick sort algorithm by following [66].

### 4.2.3 Reciprocal Rank Fusion (RRF)

Reciprocal Rank Fusion[67] calculates the rank score of a candidate term $t$ by using $t$'s ranks assigned by single rankers, as defined in (10),

$$RRF(t) = \sum_{\{r|r \in R\}} \frac{1}{k+r(t)} \qquad (10)$$

where $R$ is the set of single rankers to ensemble; $r(t)$ is the rank of $t$ given by a single ranker $r$; $k$ is a parameter used to mitigate the impact of high ranks of $t$ assigned by potential outlier systems. We set $k$ to 60 by following [67].

### 4.3 Evaluation Metrics

Precision/Recall/F-score at $n$: the averaged precision, recall and F-score at ranks 5 and 10 respectively (abbreviated as P5, R5, F5, P10, R10, and F10). Specifically, for each EHR note, the precision at rank 5 is the number of true positives in a system's top-5 predictions divided by 5 and the recall at rank 5 is the number of true positives in a system's top-5 predictions divided by the total number of positive terms in this EHR note. The F-score at rank 5 is the harmonic mean of the precision and the recall at rank 5. P5, R5 and F5 are the averaged precision, recall and F-score at rank 5 for the notes in the evaluation set. P10, R10 and F10 are calculated similarly. These metrics measure system performances for top ranks and are widely used to evaluate KE systems. We computed these metrics by using all the gold-standard important terms (including those that would never be included in the stage of candidate term extraction) as positive examples.

Area Under ROC Curve (AUC-ROC): AUC-ROC is a metrics widely used for evaluating ranking outputs. It computes the area under a ROC curve, which plots the true positive rate (y-coordinate) against the false positive rate (x-coordinate) at various threshold settings. When evaluating a system, we compute its AUC-ROC for each EHR note in the evaluation set and report the averaged value. AUC-ROC measures the performance of the global ranking. Because both candidate term extraction and ranking affect the quality of global ranking, we report two AUC-ROC metrics: $\text{AUC-ROC}_{ranking}$ and $\text{AUC-ROC}_{KE}$. $\text{AUC-ROC}_{ranking}$ is computed on the candidate terms extracted by a system. That is, if a gold-standard important term is missed in candidate term extraction, it will not be counted as a positive term when calculating the true positive rate and therefore will not affect the system's $\text{AUC-ROC}_{ranking}$. $\text{AUC-ROC}_{KE}$ is computed by using all the gold-standard important terms as positive examples and measures the combined performance of candidate term extraction and ranking.

In evaluation, we use relaxed string match to determine true positives as exact match is known to underestimate performance as perceived by human judges [73]. Specifically, we treat a term from the system output as a true positive if it either exactly matches or subsumes (e.g., "non-Hodgkin lymphoma" subsumes "lymphoma") a gold-standard important term. We allow "subsume" but not "part-of" match in relaxed string match, as previous work found that the former aligned well with human judges but the latter did not [74]. For example, a part of an important term may be too general to be important, e.g., "disease" in "Crohn's disease" and "iron" in "iron deficiency".

## 5. Results

### 5.1 Candidate Term Extraction

On average, FIT extracts 250 candidate terms per EHR note in the evaluation set, which match 89% gold-standard (physician annotated terms).

### 5.2 Evaluation Results

Our results (in Table 2) show that FIT performs significantly better than any single-view ranker (see Table 2 for *P* values).

Table 2. Performances of FIT and single rankers on the evaluation set. We report the *P* values between TOPIC and FIT. The *P* values between other single rankers and FIT are all <.001 for all metrics. A *P* value <0.05 is considered to be statistically significant.

| System | P5 | R5 | F5 | P10 | R10 | F10 | AUC-ROC$_{ranking}$ | AUC-ROC$_{KE}$ |
|---|---|---|---|---|---|---|---|---|
| FIT | **0.320** | **0.209** | **0.234** | **0.281** | **0.361** | **0.291** | **0.885** | **0.813** |
| Patient Vocabulary Model | 0.196 | 0.111 | 0.131 | 0.167 | 0.190 | 0.162 | 0.729 | 0.671 |
| TF*IDF | 0.189 | 0.115 | 0.132 | 0.156 | 0.192 | 0.158 | 0.657 | 0.604 |
| Adapted SingleRank | 0.193 | 0.130 | 0.145 | 0.160 | 0.217 | 0.171 | 0.703 | 0.640 |
| Topic Coherence | 0.251 (*P*=.006) | 0.165 (*P*=.03) | 0.185 (*P*=.01) | 0.193 (*P*<.001) | 0.259 (*P*<.001) | 0.205 (*P*<.001) | 0.784 (*P*<.001) | 0.722 (*P*<.001) |

In addition, as shown in Table 3, FIT outperforms the three ensemble methods for all metrics except having a tie with CombSum on AUC-ROC$_{KE}$. The performance difference between FIT and each baseline ensemble method is statistically significant for some metrics (see Table 3 for *P* values).

Table 3. Performances of different ensemble ranking systems on the evaluation set. We report the *P* values (if the *P* value is below 0.05) between each ensemble method and FIT. A *P* value <0.05 is considered to be statistically significant.

| System | P5 | R5 | F5 | P10 | R10 | F10 | AUC-ROC$_{ranking}$ | AUC-ROC$_{KE}$ |
|---|---|---|---|---|---|---|---|---|
| FIT | **0.320** | **0.209** | **0.234** | **0.281** | **0.361** | **0.291** | **0.885** | 0.813 |
| CombSum | 0.302 | 0.202 | 0.225 | 0.253 (*P*=.03) | 0.335 | 0.266 (*P*=.01) | 0.884 | **0.813** |
| CondorcetFuse | 0.264 | 0.168 | 0.191 | 0.218 (*P*<.001) | 0.277 (*P*<.001) | 0.225 (*P*<.001) | 0.819 (*P*<.001) | 0.753 (*P*<.001) |
| Reciprocal Rank Fusion | 0.313 | 0.208 | 0.230 | 0.249 (*P*=.004) | 0.322 (*P*=.02) | 0.260 (*P*=.01) | 0.878 (*P*=.04) | 0.807 (*P*=.04) |

Figure 3 shows the ranks (x-axis) which different NLP systems assign to the five important medical terms identified by physicians in the EHR excerpt in Figure 1, where rank 1 represents the highest rank. As shown in Figure 3, the ranks assigned by FIT are higher than those assigned by the baseline systems for most cases. The original data for Figure 3 is shown in Appendix D.

**Figure 3.** Ranks (x-axis) assigned by different NLP systems to the important medical terms in the EHR excerpt in Figure 1. Rank 1 represents the highest rank.

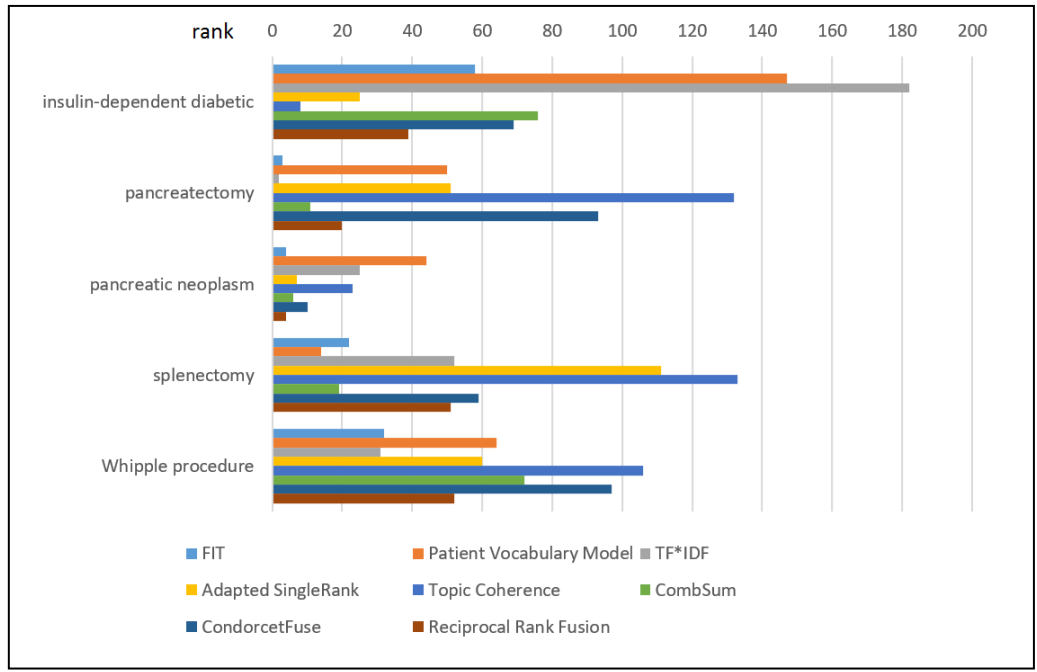

### 5.3 Impact of Parameter *d* on FIT's Performance

Figure 4 plots FIT's performance (F5, F10, AUC-ROC$_{ranking}$, and AUC-ROC$_{KE}$) for different values of the parameter *d* used in Eq. (7) (the value of each data point is provided in Appendix E). As shown in Figure 4a, when $d>0$, the F5 score is relatively stable; while the F10 score first increases, reaching a peak point at $d=0.2$, and then decreases. In addition, when $d>0$, the F5 and F10 scores at different *d* values are consistently higher than the respective F5 and F10 scores at $d=0$ (FIT equals CombSum at $d=0$). As shown in Figure 4b, the two AUC-ROC scores have the same trend, reaching the highest values at $d=0.1$ and slightly decreasing for bigger *d*'s.

**Figure 4.** Effect of *d* on FIT's ranking performance.

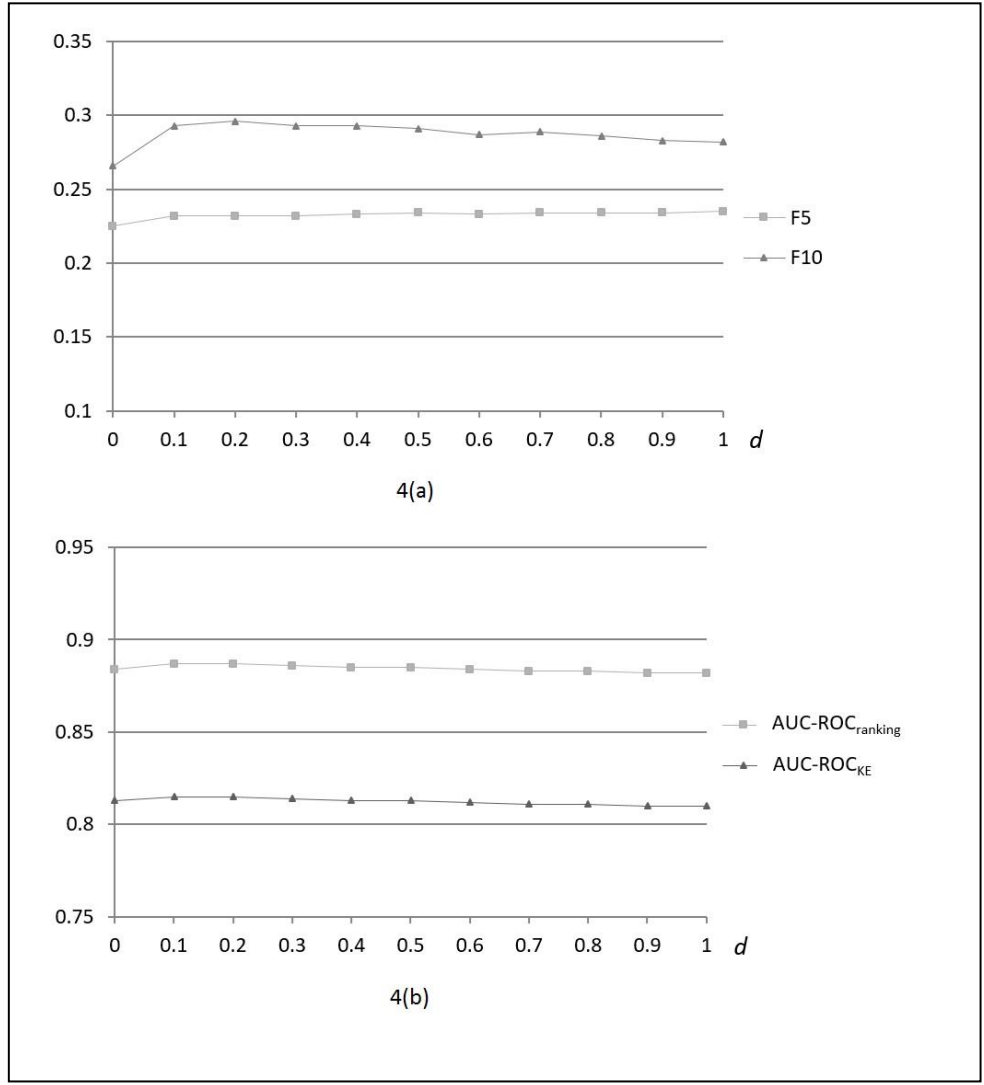

4(a)

4(b)

### 5.4 Effects of Domain-Specific Information on FIT's Performance

We compared the performances of an implementation of FIT that uses information only from TF*IDF, TextRank, and Topic Coherence (called FIT-generic) and the other three implementations that respectively add information from each of the three domain-specific knowledge resources (i.e., Patient Vocabulary Model, term unfamiliarity, and CHV-preferred semantic types) on FIT-generic.

The results (Table 4) show that adding each type of domain-specific information to FIT-generic consistently improves its performance for all metrics. The system differences are significant for AUC-ROC measures ($P<.001$).

Table 4. Performances of different implementations of FIT. FIT-generic uses only TF*IDF, TextRank, and Topic Coherence. We report the *P* values (if the *P* value is below 0.1) between each implementation and FIT-generic. A *P* value <0.05 is considered to be statistically significant.

| System | P5 | R5 | F5 | P10 | R10 | F10 | AUC-ROC$_{ranking}$ | AUC-ROC$_{KE}$ |
|---|---|---|---|---|---|---|---|---|
| FIT-generic | 0.273 | 0.192 | 0.208 | 0.232 | 0.312 | 0.244 | 0.822 | 0.755 |
| + Patient Vocabulary Model | 0.289 | 0.197 | 0.217 | 0.250 (*P*=.07) | 0.331 | 0.263 (*P*=.07) | 0.883 (*P*<.001) | 0.811 (*P*<.001) |
| + term unfamiliarity | 0.289 | 0.197 | 0.217 | 0.241 | 0.320 | 0.253 | 0.842 (*P*<.001) | 0.774 (*P*<.001) |
| + semantic type | 0.296 (*P*=.06) | 0.208 (*P*=.06) | 0.226 (*P*=.06) | 0.239 | 0.314 | 0.249 | 0.834 (*P*<.001) | 0.766 (*P*<.001) |

## 6. Discussion

### 6.1 Principle Results

Automated ranking of EHR terms based on their importance to patients is challenging because EHR notes contain abundant medical terms, among which only a small portion (4% in our case) were judged to be important by physicians. Unsupervised ranking of EHR terms is even more challenging for two reasons. First, the ranker has no supervision from annotated data. Second, and probably most important, many factors affect the importance of a term. As a result, no single criterion or type of information is sufficient to achieve adequate results in ranking candidate terms (see Rows 2-4 in Table 2). Our system FIT overcomes this problem by integrating multiple complimentary information resources. It achieves 0.885 AUC-ROC in ranking candidate terms and 0.813 AUC-ROC when counting errors from candidate term identification. This performance level is adequate, especially for unsupervised systems. Our work is an important step towards building generalizable NLP systems to facilitate personalized interventions to improve patients' EHR comprehension. FIT's output can also be used by other NLP applications including summarization and question answering.

### 6.2 FIT vs. Single Rankers

As introduced in section 3.1, unsupervised ensemble ranking methods assume the goodness and diversity of single rankers. When these conditions are compromised, an ensemble ranker is not guaranteed to outperform the best single ranker or may have mixed results across different metrics [66,68,70,75]. Our results (Table 2) show that FIT works well and outperforms each single ranker significantly for all the metrics, suggesting that our design of FIT and selection of single rankers are appropriate and effective.

Among the four single rankers, Topic Coherence performs best (Row 5 in Table 2). Topic modeling has been used to extend SingleRank to improve the state-of-the-arts in unsupervised KE [76,77]. Our results suggest that Topic Coherence, as a standalone ranker, can provide good-quality input for ensemble ranking for NLP tasks similar to KE.

SingleRank is among the state-of-the-arts and TF*IDF is frequently used as a strong baseline in unsupervised KE from scientific literature and news [33]. Both models heavily rely on word and term frequency. However, in our data, 56% of important medical terms occur only once in any individual EHR note, which partially explains why the two methods are less effective for EHRs.

Unlike other single rankers, Patient Vocabulary Model was introduced specifically to incorporate domain(task)-specific knowledge into FIT. Our results show that its performance (Row 2 in Table 2) is comparable to other single rankers and is the second best on the AUC-ROC scores. In addition, as shown in Table 4 (Row 2 vs. Row 1), adding this model on a generic FIT system (FIT-generic) that uses only general information about term importance improves performance for all metrics. These results verify the validity of our assumption behind this model, i.e., medical terms that occur in both EHRs and the CHV are important to patients for comprehending their EHRs.

### 6.3 FIT vs. Baseline Ensemble Rankers

CondorcetFuse, Reciprocal Rank Fusion, and CombSum respectively use pairwise rank order relations, rank orders, and rank scores for ensemble ranking. Previous work shows that the performances of the three methods vary for different tasks and datasets and, therefore, there is not a guaranteed winner [66,68,67,69,70,75]. Furthermore, in real-world tasks, it is likely that we have only certain types of information (e.g., pairwise rank order relations from product review) or have information from heterogeneous resources (e.g., in our case). Therefore, it is desired to have a robust ensemble ranker that can utilize different information resources flexibly.

FIT, by its design and as confirmed by our experiments, has the desired properties. For example, as shown in Table 4, FIT not only can use the complete information about rank scores and orders given by Patient Vocabulary Model to improve its performance (Row 2 vs. Row 1), but also can use the partial information about rank orders inferred from term unfamiliarity (Row 3 vs. Row 1). In addition, FIT's performance is relatively stable at different *d* values (Figure 4), confirming its robustness. These properties make FIT easily generalizable to new domains and other ranking problems.

### 6.4 Error Analysis and Future Work

We manually examined 22 notes, on which FIT has either zero recall at rank 10 or low AUC-ROC$_{KE}$ (<0.700). We identified three types of errors.

First, we used relaxed string match for evaluation but did not allow "part-of" match (for the reason discussed in section 4.3). However, in some cases, this approach underestimates the performance. For example, FIT counted it as a mistake if MetaMap recognized "invasive carcinoma" and "the colon cancer" but not "invasive carcinoma of the colon cancer", the gold-standard term.

Second, FIT depends on MetaMap, which makes mistakes. It failed to identify certain abbreviations as medical terms, e.g., A1c (a lab test for blood glucose), MRCP (Magnetic resonance cholangiopancreatography), CPPD (calcium pyrophosphate deposition disease), and TSH (a lab test for thyroid stimulating hormone). In future work, we may collect a list of common clinical abbreviations by mining a large EHR corpus and uses this list to enhance medical term identification.

Third, physicians sometimes judged common disease names (e.g., "hypertension", "diabetes", and "coronary artery disease") as important when they are the major diagnoses of a patient. These terms were frequently missed from FIT's top-10 because they were ranked low by TF*IDF (due to their high document frequencies) and the patient vocabulary model (which was trained to rank unfamiliar terms in both EHRs and the CHV high).

### 7. Conclusions

We have presented FIT, an unsupervised ensemble ranking system for identifying medical terms important to patients from individual EHR notes. FIT can combine heterogeneous information resources. It achieves promising results and outperforms benchmark unsupervised ensemble methods in ranking EHR terms. Our work is an important step towards empowering patients to comprehend their own EHR notes to improve quality of care. By using unsupervised learning and robust information fusion techniques, FIT can be readily applied to other domains and applications (e.g., document retrieval and opinion extraction).

### Conflicts of Interest
None declared.

## Acknowledgments

This work was supported in part by the Investigator Initiated Research 1I01HX001457-01 from the Health Services Research & Development Program of the United States (U.S.) Department of Veterans Affairs. We also acknowledge the support from the Institutional National Research Service Award (T32) 5T32HL120823-02 from the National Institutes of Health (NIH).
The content is solely the responsibility of the authors and does not necessarily represent the official views of the U.S. Department of Veterans Affairs, NIH or the U.S. Government.

We thank the UMassMed annotation team, including Elaine Freund, Victoria Wang, Andrew Hsu, Barinder Hansra and Sonali Harchandani, for creating the evaluation corpus and thank Weisong Liu for technical support in collecting EHR notes. We also thank the anonymous reviewers for their constructive comments and suggestions.

## Appendix A. Frequency distribution of semantic type in Consumer Health Vocabulary

Figure A.1. Frequency distribution of semantic type in Consumer Health Vocabulary. x-axis: rank of semantic types by frequency. Rank 1 represents the most frequent semantic type.

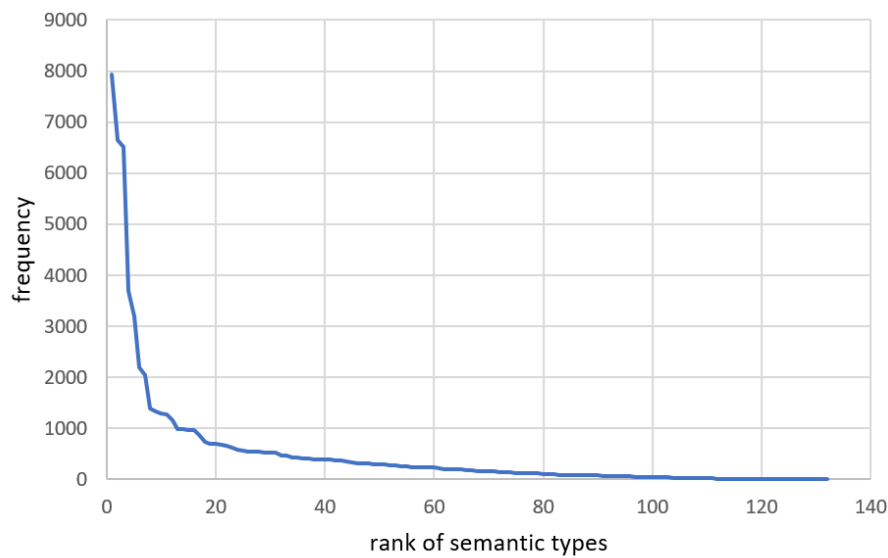

# Appendix B. Twelve high-frequency UMLS semantic types in Consumer Health Vocabulary used to prioritize important medical terms in EHR notes

| UMLS semantic type | Example EHR terms |
| --- | --- |
| Pharmacologic substance | Advair, Budesonide, insulin, NSAIDs, Spironolactone |
| Disease or syndrome | autoimmune hemolytic anemia, gastroesophageal reflux, pancytopenia, Sjogren's syndrome, osteoporosis |
| Organic chemical | Atenolol, Vincristine, Warfarin, Wellbutrin, Zocor |
| Finding | alopecia, hematuria, hypertension, NSTEMI (Non-ST-elevation myocardial infarction), retinopathy |
| Therapeutic or preventive procedure | chemotherapy, dialysis, immunosuppression, kidney transplantation, pancreatectomy |
| Amino acid, peptide, or protein[2] | basal insulin, Rituxan, Neupogen, Synthroid, hemoglobin A1C, HPL (human placental lactogen) |
| Body part, organ, or organ component | adrenal glands, coronary arteries, cranial nerves, lymph nodes, thyroid nodule |
| Sign or symptom | lower extremity edema, sciatica, scleral icterus, syncopal episodes, vertigo |
| Medical device | Foley catheter, defibrillator, insulin pump, pacemaker, pedometer |
| Neoplastic process | dermoid, large B cell lymphoma, pancreatic neoplasm, thyroid nodule |
| Injury or poisoning | bruising, distal radial fracture, exposure to asbestos, spinal compression fractures, Methotrexate toxicity |
| Laboratory procedure | hepatitis B serology, LFTs, lipid panel, sedimentation rate, urinalysis |

---

[2] EHR terms in this topic split into two subtopics: medicine (denoted by their ingredients) and laboratory measure.

# Appendix C. Guidelines for annotating medical terms important to patients in EHR notes

1. Goal/task: identifying at least five most important medical terms per EHR note which the patients need to know in order to better understand their EHR notes

In general, the goal can be achieved by selecting the minimum number of medical terms, which if the patients know, they will have a significant understanding of their clinical diseases and symptoms without being overwhelmed.

We provide operational rules in Section 2 to help achieve this goal.

2. Selection criteria
(1) Include terms that represent the main concept of each EHR note
Note: The most important medical terms that patients should know shall be straight-forward clinical knowledge, rather than complex clinical knowledge that may confuse patients or may need additional explanation

(2) Include terms that are related to the main concepts identified in (1) and can help patients' comprehension of the most important clinical concepts in their EHR notes

Note: These related terms also shall be straight-forward clinical knowledge, rather than complex clinical knowledge that may confuse patients or may need additional explanation

-------------------------------------------------------------------------------------------------
Notes:

The annotation guideline was developed by a group of physicians and non-physicians.

Textbox 1.1 shows an example annotation to illustrate "complex clinical knowledge that may confuse patients or may need additional explanation". The physicians did not annotate the terms "left axis deviation" and "nonspecific t-wave lowering", whose comprehension requires complex knowledge about EKG curve and heart function.

Textbox 1.1 An example annotation where terms judged by physicians to be important to patients are bracketed.

> Today's *EKG* shows <atrial fibrillation> with *controlled ventricular response*, *left axis deviation*, *nonspecific t-wave lowering* in many *leads*, and ongoing *fluctuation* since last year and the year before.

**Appendix D. Ranks assigned by different systems to the five important medical terms in the EHR excerpt in Figure 1**

| Systems | insulin-dependent diabetic | pancreatectomy | pancreatic neoplasm | splenectomy | Whipple procedure |
|---|---|---|---|---|---|
| FIT | 58 | 3 | 4 | 22 | 32 |
| Patient Vocabulary Model | 147 | 50 | 44 | 14 | 64 |
| TF*IDF | 182 | 2 | 25 | 52 | 31 |
| Adapted SingleRank | 25 | 51 | 7 | 111 | 60 |
| Topic Coherence | 8 | 132 | 23 | 133 | 106 |
| CombSum | 76 | 11 | 6 | 19 | 72 |
| CondorcetFuse | 69 | 93 | 10 | 59 | 97 |
| Reciprocal Rank Fusion | 39 | 20 | 4 | 51 | 52 |

## Appendix E. Effect of *d* on FIT's ranking performance

| *d* | P5 | R5 | F5 | P10 | R10 | F10 | AUC-ROC$_{ranking}$ | AUC-ROC$_{KE}$ |
|---|---|---|---|---|---|---|---|---|
| 0 (CombSum) | 0.302 | 0.202 | 0.225 | 0.253 | 0.335 | 0.266 | 0.884 | 0.813 |
| 0.1 | 0.318 | 0.206 | 0.232 | 0.281 | **0.366** | 0.293 | **0.887** | **0.815** |
| 0.2 | 0.318 | 0.208 | 0.232 | 0.287 | 0.365 | **0.296** | **0.887** | **0.815** |
| 0.3 | 0.316 | 0.209 | 0.232 | 0.283 | 0.362 | 0.293 | 0.886 | 0.814 |
| 0.4 | 0.318 | 0.208 | 0.233 | 0.283 | 0.364 | 0.293 | 0.885 | 0.813 |
| 0.5 | 0.320 | **0.209** | 0.234 | 0.281 | 0.361 | 0.291 | 0.885 | 0.813 |
| 0.6 | 0.320 | 0.208 | 0.233 | 0.278 | 0.355 | 0.287 | 0.884 | 0.812 |
| 0.7 | 0.322 | **0.209** | 0.234 | 0.280 | 0.357 | 0.289 | 0.883 | 0.811 |
| 0.8 | 0.322 | **0.209** | 0.234 | 0.278 | 0.351 | 0.286 | 0.883 | 0.811 |
| 0.9 | 0.322 | **0.209** | 0.234 | 0.276 | 0.348 | 0.283 | 0.882 | 0.810 |
| 1.0 | **0.324** | **0.209** | **0.235** | 0.274 | 0.346 | 0.282 | 0.882 | 0.810 |